# AI Uncertainty Based on Rademacher Complexity and Shannon Entropy

Mingyong Zhou
School of Computer Science and
Communication Engineering
GXUST, Liuzhou, China
Zed6641@hotmail.com

*Abstract*—**In this paper from communication channel coding perspective we are able to present both a theoretical and practical discussion of AI's uncertainty, capacity and evolution for pattern classification based on the classical Rademacher complexity and Shannon entropy. First AI capacity is defined as in communication channels. It is shown qualitatively that the classical Rademacher complexity and Shannon entropy used in communication theory is closely related by their definitions, given a pattern classification problem with a complexity measured by Rademacher complexity. Secondly based on the Shannon mathematical theory on communication coding, we derive several sufficient and necessary conditions for an AI's error rate approaching zero in classifications problems. A 1/2 criteria on Shannon entropy is derived in this paper so that error rate can approach zero or is zero for AI pattern classification problems. Last but not least, we show our analysis and theory by providing examples of AI pattern classifications with error rate approaching zero or being zero.**

*Impact Statement:* **Error rate control of AI pattern classification is crucial in many life related AI applications. AI uncertainty, capacity and evolution are investigated in this paper. Sufficient/necessary conditions for AI's error rate approaching zero are derived based on Shannon's communication coding theory. Zero error rate and zero error rate approaching AI design methodology for pattern classifications are illustrated using Shannon's coding theory. Our method shows how to control the error rate of AI, how to measure the capacity of AI and how to evolve AI into higher levels.**

## I. INTRODUCTION

Artificial Intelligence (AI) based on deep learning neuron networks are widely used in CT imaging, pattern recognitions, medical diagnosis, and scientific computing. Usually large neuronal network architectures are first selected for a specific purpose and the neuron network are trained by large samples of the data from the specific applications such as CT imaging for medical diagnosis. Secondly fully trained neuron networks are used as AI model to diagnose unknown patterns for example CT images to be normal or abnormal ones. Current AI model depends upon the large samples training methods and control the error rate as small as possible. However in practice one wishes to control the error rate to be approaching zero or even to be zero in many life related applications such as in medical diagnosis. It is therefore an important task for AI designers to design an AI model that can control the error rate in an efficient way so that AI model can be used in life related crucial applications. There are huge references in analyzing upper and lower bounds for neuron network errors in accurately recognizing the patterns. However very few papers addressed AI design methods to control the error rates to be zero or approaching zero as much as desired pplications. In this article, we are investigating in this direction by providing both a theory and a pragmatic methodology.

## II. BACKGROUND

As Artificial Intelligence (AI) is based on trained various neuron networks such as CNN or RNN and becomes widespreading and applied in various industries, AI uncertainty investigations based on various neuron networks obviously becomes more and more important in practical applications in that probability errors from data driven AI model itself and errors from nuisance or noise in engineering applications are inevitable in practice. Given a problem with certain complexity measured by different measures such as Rademacher complexity and Vapnik-Cheronenkis (VC) dimensions etc, it is desired in practice that one can design a neuron network trained by deep learning algorithms so that the trained neuron network can accurately classify patterns with no errors or even small errors. It is therefore our wish that after deep learning training, neuron network can classify the test patterns with no errors or with errors approaching zero. Errors of two layer forward neuron networks for classification problems and approaching functions were theoretically and in quantity investigated in [1] and [2]. Upper and lower bounds depending upon various parameters of neuron networks are derived. In this paper, however, by applying Shannon's mathematical theory on communications channel coding [3], we address the pattern classification errors of trained neuron networks from Shannon's entropy perspective. We provide a generalized framework to discuss AI uncertainty, capacity and evolution with the existence of probability errors from data driven AI models[1][2] and errors from nuisance and noise in AI applications[3]. We prove that a sufficient and necessary condition that ensures classification error of trained neuron network can approach zero as much as desired. Other sufficient conditions are derived as well. We do not limit our neuron network structures into specific CNN or RNN or two layer forward neuron networks, but provide a more generalized discussion on the errors introduced at trained neuron network phase and at application phase with nuisance and noise. Learning algorithms are therefore not included in this letter paper.

In **Section III**, a methodology is described on how to design an AI system that can ensure the zero error approaching purpose in AI systems by using Shannon's channel coding theory. **Section IV** is devoted to the theory and analysis based on Shannon's entropy on communication channel capacity and classical Rademacher complexity. We derive several sufficient and necessary conditions regarding classification error rate by observing a qualitatively relation between Rademacher complexity and Shannon entropy. Analysis are done in this section and sufficient /necessary conditions are derived for zero error approaching AI system. Examples are given for zero error approaching AI system as well. This paper is concluded with **section V** in which major results are summarized and future work direction is indicated..

## III. METHODOLOGY

Considering a scenario in which a clinic is doing CT scanning and then needs an AI to process and identify COVID-19 lung images from normal ones per second. We wish to design an AI to classify the normal images and COVID-19 images with error approaching zero or being zero during a given interval time. We first summarize our methodology as in the following steps and then explain our AI design methodology. The principle is to apply Shannon channel coding theory to ensure zero error approaching property of designed AI system. As illustrated in Figure 1,

1) AI core is designed with a proper neuron network architecture such as CNN, RNN or forward 2 layer neuron network with each neuron with sigmoid functions property. ERROR in figure1 denotes the system error due to training.NOISE denotes the nuisance in real applications.

2) A training is completed with Back Propagation(BP) algorithms with large known samples such as CT imaging samples for COVID19 lung images and normal lung images. There we have 2 types of patterns, one is COVID19 lung CT images and the other pattern is normal lung images without COVID19 virus affections. Note that there is an error rate (denoted by ERROR in figure 1) in this AI core in that normal lung images may be classified into COVID19 case.

3) Given ERROR obtained at 2), if ERROR satisfies conditions which are to be discussed in the following section, a simple or advanced coding method exists so that the images can be classified with error approaching zero.

4) For identified failure images, go to 2) and train the failed samples so that AI evolves to a higher level.

5) The process is completed until there are no more failed images. This process can be thought as an adaptive AI methodology.

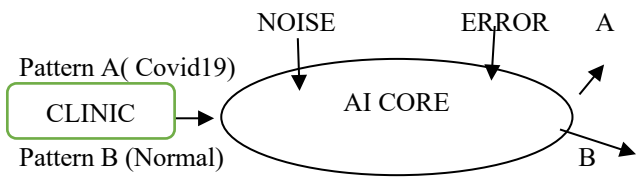

Figure1: AI system with ERROR and NOISE

## IV. THEORY AND ANALYSIS

In this section, we first relate neuron network based AI system to a communication channel and then derive two theorems regarding errors based on the quality relations between Rademacher complexity and Shannon entropy. We first overview the main results of Shannon theory by reviewing Shannon's definition on channel/AI capacity and information rate. Then we present two theorems for an AI model for pattern classification problems. We will show that AI and communication channel resemble in principle. Further we elucidate a quality relation between classical Rademacher complexity and Shannon entropy from their definitions. First the following Shannon condition are well known to ensure a proper encoding for zero error approaching communication channels.

$$R < C$$
$$C = Max(H(x) - H_y(x)) \qquad (1)$$

where $C$ is defined as the channel/AI capacity in Shannon theory for a noisy channel, and $R$ is transmission rate. *Max* denotes maximum value and $H(x)$ denotes Shannon entropy and $H_y(x)$ denotes Shannon conditional entropy [3]. Condition (1) by Shannon theory indicates that the transmission rate $R$ must not exceed channel/AI capacity $C$ in order to recover the original information by a proper coding with errors approaching zero[3]. Now with an AI model for classification problems, it is natural to relate AI system to a communication channel with the following confronted problem: given a pattern classification problem with known complexity measured by a known Rademacher complexity, to what extend a neuron network should be trained so that AI deep learning neural network with $H_y(x)$ has the ability to recognize the correct patterns with errors approaching to zero? In AI system, $H_y(x)$ can be interpreted as the total errors introduced by the trained AI model [1][2] and the errors introduced by inevitable nuisance and physical noises in real AI applications. We wish to explore the conditions that an AI system can allow such total errors and how to reach this goal, given a complex classification problem. In the following, we will show that based on Rademacher complexity, Shannon entropy and Shannon communication coding theory, sufficient/necessary conditions can be found for zero error approaching AI system.

Obviously $R$ transmission rate in Shannon theory is closely related to Rademacher complexity for a given complexity problem. To show this, we only need to note the definition of classical Rademacher complexity $R(F)$ by equation (2), where $F$ denotes the hypothesis function space of problems. Rademacher complexity $R(F)$ *thus* reaches maximum value of 1.0 in case that the most complex binary +1, -1 are matched exactly. For simplicity we use $\phi$ to denote the mapping from Rademacher complexity $R(F)$ in equation (2) to Shannon rate $R$ in equation (1). A quality relationship of $\phi$ is shown in Figure2 where horizontal axis denotes Rademacher $R(F)$ and vertical axis denotes Shannon rate $R$ in Eq.(1). To show 2 examples, Rademacher complexity of zero implies that only ONE fixed 1, or -1 is transmitted. Therefore Shannon rate is 1. Shannon rate of zero on the other hand implies that infinite 1 or 0's are transmitted during an interval time. Therefore Rademacher complexity is 1.

$$R(F) = E\{\frac{1}{n}\sup_{f \in F}[\sigma_i \times f(X_i)]\},$$
$$\sigma_i = \pm 1, f(X_i) = \pm 1$$

(2)

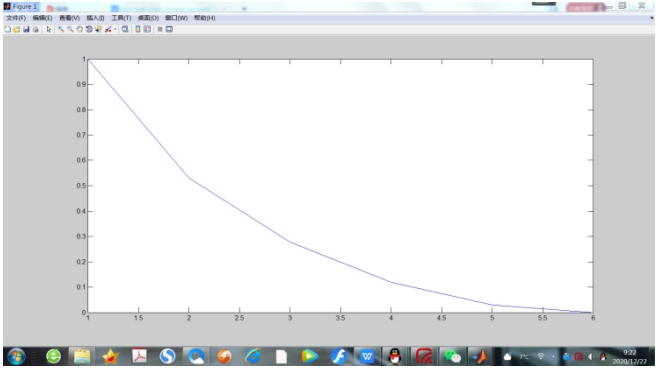

Figure 2: Quality relationship $\phi$ between Shannon rate $R$ in Eq.(1) and Rademacher $R(F)$ in Eq.(2)

Based on the above descriptions, we now present two conditions as follows

$$A), \phi\{R(F)\} \leq \frac{1}{2} Max(H(x))$$

(3)

Condition A) of equation (3) indicates a condition of Rademacher complexity.

$$B), Min(H_y(x)) \leq \frac{1}{2} Max(H(x))$$

(4)

Condition B) of equation (4) indicates a condition regarding Shannon conditional entropy for errors.

**Theorem 1:** Shannon condition (1) is a sufficient condition for conditions (3) and (4) but is not a necessary condition for (3) and (4) generally.

**The proof** is a case analysis starting from equation (1). By Shannon theory on the noise channel with conditional entropy $H_y(x)$, to ensure transmit with no errors starting from equation (1), the following (5) must be satisfied

$$R < Max(H(x)) - \min H_y(x)$$

(5)

Equation (5) implies that

$$R + \min H_y(x) < Max(H(x)),$$

(6)

Taking into considerations of quality mapping $\phi$ as in Figure 2, one has

$$\phi\{R(F)\} + Min(H_y(x)) < Max(H(x))$$

(7)

The analysis is now performed as part A) and part B):

A) , if $\phi\{R(F)\} \geq \min H_y(x)$ then

$$2\min(H_y(x)) \leq \min(H_y(x)) + \phi\{R(F)\} \leq Max(H(x))$$

$$\min(H_y(x)) \leq \frac{1}{2} Max(H(x))$$

(8)

B) , if $\phi\{R(F)\} \leq \min H_y(x)$, then

$$\phi\{R(F)\} \leq \min(H_y(x)) \leq Max(H(x)) - \phi\{R(F)\}$$

$$\phi\{R(F)\} \leq \frac{1}{2} Max(H(x))$$

(9)

□

**Theorem 2:** Under conditions

$$\phi\{R(F)\} \geq Min(H_y(x)), R \leq 1/2 Max(H(x)) \quad (10)$$

Shannon condition (1) is a both sufficient and necessary condition for condition (4).

**The necessary proof** is a direct derivation as shown in the following inequality
$$C = Max(H(x)) - Min(H_y(x)) > 1/2 Max(H(x)) > R$$

(11)

**The sufficient proof** is included in the proof of theorem 1 under the condition indicated.

□

Theorem 2 has a significant importance in practical use.

Without loss of generality we can assume that Max(H(x))=1 in the subsequent discussions.

The physical interpretation is that given the condition of R<1/2Max(H(x)) which means the Rademacher complexity is of above moderate complexity as shown in figure 2, as long as condition (4) and $\phi\{R(F)\} \geq Min(H_y(x))$ are satisfied, Shannon condition of (1) is thus satisfied. In other words, errors by a way of proper coding approach zero as much as desired according to Shannon theory [3]. Due to 1/2 co-efficiency is involved in the conditions, we terms 1/2 criteria in this paper. To demonstrate 2 extreme examples, for a most complex problem,

$$R = 0 < \frac{1}{2}, \phi\{R(F) = 1.0\} = 0 \geq Min(H_y(x)) \quad (12)$$

Therefore total errors introduced by $H_y(x)$ must be zero, which means there is zero error tolerance for most complex

problems. For a moderate complex problem $R(F)<1.0$, therefore

$$\forall \varepsilon, 0 < \varepsilon \leq \frac{1}{2}, R = \varepsilon \leq \frac{1}{2}, \phi\{R(F)\} = \varepsilon \geq Min(H_y(x)) \quad (13)$$

Thus for a given moderate complex problem, there is an error tolerance for AI system so that a zero error approaching is possible by a coding method. Theorem 2 indicates that for a given moderate complexity problem, AI system total errors must be controlled strictly so that a zero error approaching communication is possible by a proper coding.

## V. EXAMPLES AND FUTURE WORK

Suppose we are faced with classifying two patterns denoted by x that denotes normal lung images and y that denotes COVID-19 lung images per second by a trained AI system. In this section, we provide one simple example and one complex example to illustarte our theory and analysis in section IV. One simplest "coding" method is by repeating the patterns by a certain iterations of numbers that depends upon ERROR of AI core in Figure 1 under condition that equation (10) is satisfied. A more complex coding method is, for example, a Muller coding. By theorem 2, Shannon conditional entropy $H_y(x)$ of AI total error must be no more than 1/2. Therefore total error rates by trained AI model in this case must be less than 12% according to Shannon's entropy [3]. Once the total errors are bounded, we can design a coding method to ensure a zero error approaching AI system in applications. However if condition (10) is not satisfied, zero error approaching AI system is NOT possible anyhow.

In this paper Rademacher complexity is related to Shannon's entropy. Based on the derived results in this paper, we can as well extend to other complexity measures such as Vapnik-Cheronenkis (VC) and Gaussian complexity by provoking the results in [6]. Criteria and condition based on other complexity measures are possible.

## IV. CONCLUSIONS

In this paper, we study AI capacity, uncertainty and evolution from communication channel coding perspective. A methodology framework based on communication channel coding insight is proposed to investigate AI's capacity and uncertainty as well as AI's evolution. First classical Rademacher complexity is outlined and its relations with Shannon entropy is elucidated. Secondly by comparing AI system to a communication channel, we are able to apply Shannon theory and entropy to propose a zero error approaching AI methodology. A 1/2 criteria is derived from Shannon theory for zero error approaching AI. Several sufficient/necessary conditions are derived for zero error approaching AI. The theoretical results derived in this paper can provide an insight into zero error approaching AI framework. Methodology and example are shown on how to evolve AI to reach lower error rates and how to utilize existing simple or complex coding methods to ensure zero error or zero approaching purpose in applications. The main results are summarized as follows:

(1), Classical Rademacher complexity is closely related to Shannon rate. Their qualitative relations are demonstrated.

(2), For a pattern classification problem with moderate Rademacher complexity, a sufficient and necessary condition is derived to ensure a zero error or zero approaching error for AI.

(3), A pragmatic insight is provided for AI deep learning neural network, that is, the higher Rademacher complexity of given problems, the less tolerant for AI mistakes in order to reach zero approaching errors. However for problems with very high Rademacher complexity, if AI is evolving in a sense that AI can reduce own errors and mistakes, zero approaching error rates are possible for AI in the long run.